# An attention-gated convolutional neural network for sentence classification

Yang Liu [a,*], Lixin Ji [a], Ruiyang Huang [a], Tuosiyu Ming [a], Chao Gao [a] and Jianpeng Zhang [a, b]

[a] *National Digital Switching System Engineering and Technological R&D Center, Zhengzhou, China;*
[b] *Eindhoven University of Technology, 5600 MB Eindhoven, The Netherlands;*
33603@bit.edu.cn (Y.L.); jlxndsc@139.com (L.J.); 18337176095@139.com (R.H.); 1139446336@qq.com (T.M.); chaosndsc@163.com (C.G.); j.zhang.4@tue.nl (J.Z.)

**Abstract.** The classification of sentences is very challenging, since sentences contain the limited contextual information. In this paper, we proposed an Attention-Gated Convolutional Neural Network (AGCNN) for sentence classification, which generates attention weights from the feature's context windows of different sizes by using specialized convolution encoders. It makes full use of limited contextual information to extract and enhance the influence of important features in predicting the sentence's category. Experimental results demonstrated that our model can achieve up to 3.1% higher accuracy than standard CNN models, and gain competitive results over the baselines on four out of the six tasks. Besides, we designed an activation function, namely, Natural Logarithm rescaled Rectified Linear Unit (NLReLU). Experiments showed that NLReLU can outperform ReLU and is comparable to other well-known activation functions on AGCNN.

## 1. Introduction

Recently, text classification has drawn much attention from researchers. It is one of the essential tasks of the natural language processing and has been widely studied in various communities such as text mining [1] and information retrieval [14]. The primary objective of text classification is to assign texts to predefined categories according to its content automatically. The text is a typical kind of unstructured information and is complicated for the computer to understand and handle. Meanwhile, due to the limited semantic information of the sentence text carriers, it cannot provide enough word co-occurrence statistics or other features for calculating the similarity of text [46]. Therefore, sentence-level classification task has become a challenging problem.

The representation methods of the text have an enormous impact on the performance of classifiers. Traditional classification approaches mainly use the representations of statistical indicators of words such as Term Frequency-Inverse Document Frequency (TF-IDF) [19,50]. Such weighting schemes do not take the positional factors of the feature words into consideration and the feature words that have larger weights are usually not representative. Some classification algorithms such as Naive Bayes (NB) [35] and Support Vector Machine (SVM) [22] use the bag-of-words [15] for feature extraction. However, these approaches will encounter problems with data sparsification when the training set is small. Besides, there are also approaches based on n-grams including Naive Bayes SVM (NBSVM) [49], Multinomial Naive Bayes (MNB) [49] and $SVM_S$ [42]. However, by using these traditional methods, as the amount of training data increases, the dimension of feature vector will increase or even explode. And not all features are important for classifying, removing the features that are redundant or misleading is hard. Also, when the length of the sentence is short, the limited statistical information makes it difficult for traditional methods to construct effective text representation.

---

*Corresponding Author: Yang Liu, National Digital Switching System Engineering and Technological R&D Center, Zhengzhou, China. E-mail: 33603@bit.edu.cn.

To resolve these problems, distributed representation [47,32,36] was introduced in natural language processing systems, thereby promoting the development of the neural network-based approaches. Distributed representation can map words into dense real-valued and low-dimensional word vectors. Word vectors can obtain semantic and syntactic information through an unsupervised learning on corpus [37]. The sentence embedding representation can be obtained while the word vectors are learned [27]. The advantage of distributed representation is that there is no need for feature filtering or transformation.

Regarding the sentence classification based on distributed representation, neural networks-based methods are widely studied. Dominant methods in recent years are Recurrent Neural Networks (RNNs, particularly Long Short-Term Memory networks, LSTMs) and Convolutional Neural Networks (CNNs). Researchers make full use of the distributed representation of sentences and their limited context information by constructing different and sophisticated network structures. LSTMs not only learn the information at the current moment but also in the previous sequence, which is very suitable for processing text sequences. By generalizing the LSTM model to the tree-structured network topologies (Tree-LSTM) [48] or regularizing the LSTM by linguistic knowledge (LR-LSTM) [41], the network can exploit the grammatical information of the sentence. In most previous work, the supervised learning based on the single task often suffer from insufficient training data. Liu et al. [30] proposed a method for learning recurrent neural models based on a multi-task learning framework. By introducing the dynamic compositional neural networks over the tree structure [31], Tree-LSTM could capture the richness of the compositionality of sentences. However, the complex architecture of LSTMs suggests that they may not be the most efficient network structure [13].

CNNs are more computationally efficient because they make full use of the parallelism of Graphics Processing Units (GPUs), and they process information in a layered manner, which makes them easier to capture complex relationships in sentences. Blunsom et al. [3] proposed a dynamic *k*-max pooling operation, which was applied in CNN for the semantic modeling of sentences. This operation is an alternative to the parse tree, but its scheme is sophisticated. Kim [24] reported a series of experiments with simple CNN structures for sentence classification, and achieve better results in four out of the seven tasks. They found that the performance of the model is significantly improved on the pre-trained word vectors. There are also attempts to use one-hot representation as input to preserve the information of word order [23], but the sentences are too brief for such high dimensional encoding to provide sufficient information. Character-level vectors can also be used as input for CNN [54], but only for text where words are composed of characters. It is well known that in the computer vision field, the deeper the network, the better the performance should be [18].

Conneau et al. [4] also attempted towards this goal and presented a very deep CNN architecture which operates directly at the character level. However, the performance on the sentence-level classification tasks seems not reasonable. In addition, applying data augmentation methods could also improve the performance of the model [52]. Zhao et al. [53] investigated the capsule networks with dynamic routing for sentence classification.

Traditional CNN-based methods for sentence classification intend to use the pooling layer to find the significant abstract features of the words or segments that could most likely help produce the prediction result correctly. However, the mechanism of which features should be large enough to affect the prediction results is still unclear. Distributional hypothesis [21] considers that a word is characterized by the company it keeps. Similarly, we believe that context is the key to control and discriminate the influence extent of the words' or segments' features. In this paper, we construct an attention-gated layer before pooling layer for generating attention weights from feature's context windows by using specialized convolution encoders, to control the influence of target word or segment features. The attention-gated layer could help the pooling layer to find the genuinely important features. We denote our model as Attention-Gated Convolutional Neural Network (AGCNN). Experimental results demonstrated the effectiveness of our method. Besides, our proposed activation function (i.e., Natural Logarithm rescaled Rectified Linear Unit, NLReLU) is found to be comparable to other well-known activation functions.

In summary, the main contributions of this work are as follows:

- We propose a new CNN model, AGCNN, for sentence classification. The benchmark and visualization experiments demonstrate the effectiveness of our new model.
- We propose an activation function, namely, NLReLU. Experimental results show that NLReLU could outperform ReLU and is comparable to several well-known activation functions on AGCNN.
- Empirical results on six sentence classification tasks demonstrate that our model is able to achieve up to 3.1% higher accuracy than standard CNN models, and gain competitive results over the 13 strong baselines on four out of the six tasks.

The remainder of this paper is organized as follows. Section 2 gives an overview of related work. Section 3 presents the proposed model in details. Section 4 reports and discusses our experimental results. Finally, we draw our conclusions in Section 5.

## 2. Related work

LSTM [44] introduced the gating mechanism to the RNN [8], which enables RNN to remove or add information to the state of the cell. A gating mechanism usually consists of a network layer that has passed the Sigmoid [10] activation unit (generating the gating weights) and a multiplication operation. The gating weights, which usually values within the interval [0,1] (where 0 and 1 mean wholly discarded and reserved, respectively), limit the amount of information that can pass through the gates. With well-designed gating mechanisms, LSTM could learn longer range dependencies and effectively alleviate the gradient vanishing or exploding problem. Gated Convolutional

Neural Network (GCNN) [7] firstly introduced the gating mechanism into CNN for the language modeling, which could reduce the vanishing gradient problem for deep architectures. GCNN [7] utilized half of the abstract features as the gating weights to control the other half abstract features. However, since the weights and the abstract features are convolved at the same level, the information carried by the control weights is very monotonous. In this paper, we also introduce the gating mechanism in CNN, but the control weights, i.e., attention weights, are generated by a variety of specialized convolution kernels. Therefore, the contextual information of a particular context window is integrated into the control weights.

Attention mechanisms attempt to mimic the human's perception, which focus attention selectively on parts of the target areas to obtain more details of the targets while suppressing other useless information. Mnih et al. [33] firstly applied attention mechanism in RNN for image classification. Then the extensions of the attention-based RNN model are applied to various NLP tasks [2,28]. Attention mechanism in neural networks has attracted much attention and has been applied in a variety of neural network architectures including encoder-decoder [55]. The process of focusing attention in these architectures mainly reflected in the calculation of the weight coefficient. The larger the weight, the more the attention focused on its corresponding value, that is, weight represents the importance of information, and value is its corresponding information. Recently, how to use the attention mechanism in CNNs has become a research hotspot [51].

Activation functions have a crucial impact on the neural networks' performance. Sigmoid [10], Rectified Linear Unit (ReLU) [38], Softplus [38], Leaky ReLU (LReLU) [34], Parametric ReLU (PReLU) [17], Exponential Linear Unit (ELU) [5] and Scaled Exponential Linear Unit (SELU) [26] are all fairly-known and widely-used activation units. Activation functions make it possible to carry out the non-linear transformation of the input to solve the complex problems. However, it may also bring with disadvantages, e.g., vanishing gradient and neuronal death. Therefore, it is essential to choose the appropriate activation function for the neural network.

## 3. The proposed model

CNN is very suitable for natural language processing, because CNN not only allows to precisely control the length of dependencies but also enables nearby input elements to interact at lower layers while distant elements interact at higher layers, and CNN can produce the hierarchical abstract representations of the input text by stacking multiple convolution layers. Most current methods for sentence classification based on CNN intend to utilize the pooling layer to find the most significant features. In this paper, we construct an attention-gated layer before pooling layer to identify critical features, suppress the impact of other unimportant features and help pooling layer find the genuinely crucial features.

In this section, we describe our model in detail. As depicts in Fig. 1, our model consists of a convolutional layer operating on the input sentence matrix, an attention-gated layer, a max-over-time pooling layer, and a fully connected layer with dropout and softmax output. We choose a sentence with the length $n$ of 7 and the word vectors' dimensionality $d$ of 4 as an example. In the demo model shown in Fig. 1, the first convolutional layer uses convolution kernels with the window size $h$ of 2 or 3 words, and the convolution layer in the attention gated layer uses convolution kernels with the window size $k$ of 1 or 3 features. Let $e_i \in \mathbb{R}^d$ be the $d$-dimensional word vector corresponding to the $i$-th word in the sentence. An input sentence embedding matrix of sentence with length $n$ (padded when necessary) is represented as

$$E_{1:n} = [e_1, e_2, \cdots, e_n]^T \tag{1}$$

where $E_{1:n} \in \mathbb{R}^{n \times d}$.

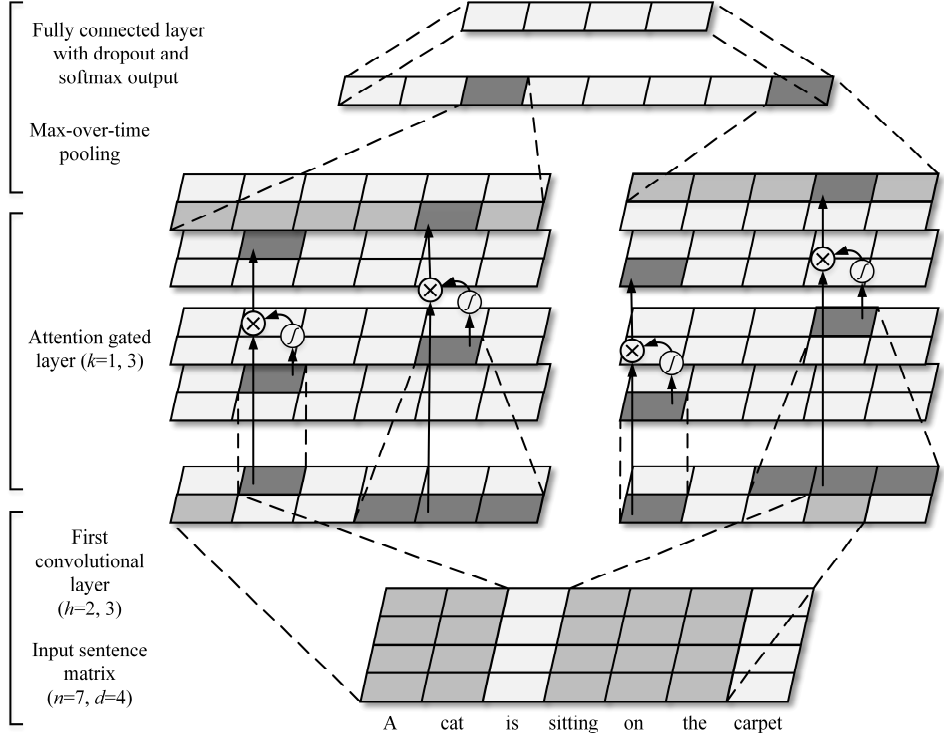

Fig. 1. The illustration of AGCNN with input sentence length $n = 7$.

In the first convolutional layer, a convolution kernel $W \in \mathbb{R}^{h \times d}$ is applied to a window of $h$ words to produce a new feature. An abstract feature $c_i \in \mathbb{R}$ is generated from a window of words $E_{i:i+h-1}$ by

$$c_i = f\left(g\left(W \otimes E_{i:i+h-1}\right) + b\right) \tag{2}$$

where $\otimes$ is the element-wise product between matrices, $b \in \mathbb{R}$ is a bias term, function $g(\cdot)$ sums up all the elements of a matrix and $f(\cdot)$ is a non-linear activation function. This convolution kernel will slide on the input sentence matrix and is applied to each possible window of words in the sentence $\{E_{1:h}, E_{2:h+1}, \cdots, E_{n-h+1:n}\}$ to produce a feature map

$$C = [c_1, c_2, \cdots, c_{n-h+1}]^T \qquad (3)$$

with $C \in \mathbb{R}^{(n-h+1)\times 1}$. Multiple kernels vary from different window sizes are applied to obtain multiple feature maps. Each feature map $C$ is then fed into the attention-gated layer.

The attention-gated layer consists of a convolutional layer and a gating mechanism. In this convolutional layer, a convolution kernel $V \in \mathbb{R}^{k\times 1}$ is applied to all the context features with window size $k$ (padded when necessary) of every feature $c_j\ (j = 1, 2, \cdots, n-h+1)$ in feature map $C$ to produce the attention weight matrix

$$A = [a_1, a_2, \cdots, a_{n-h+1}]^T \qquad (4)$$

**Theorem 1.** Each attention weight element $a_j (j = 1, \cdots, n-h+1)$ in matrix $A$ is computed by equation

$$a_j = \begin{cases} g\left(V \otimes C_{j-\frac{k-1}{2}:j+\frac{k-1}{2}}\right) & (j = 1, \cdots, n-h+1,\ \text{when}\ k\ \text{is odd}) \\ g\left(V \otimes C_{j-\frac{k}{2}+1:j+\frac{k}{2}}\right) & (j = 1, \cdots, n-h+1,\ \text{when}\ k\ \text{is even}) \end{cases} \qquad (5)$$

where $a_j \in \mathbb{R}$ and $A \in \mathbb{R}^{(n-h+1)\times 1}$. $C_{j-\frac{k-1}{2}:j+\frac{k-1}{2}}$ and $C_{j-\frac{k}{2}+1:j+\frac{k}{2}}$ is the context features of $c_j$.

*Proof.* For convolution (1D convolution as an example) with stride $s$ of 1, input height of $H$, and the height of the kernel of $k$, the number of the elements that need to be padded on the input is

$$\begin{aligned} p_{need} &= (\frac{H}{s} - 1) \times s + k - H \\ &= k - 1 \end{aligned} \qquad (6)$$

The number of elements that need to be padded up and down on the input is

$$\begin{cases} p_{top} = \dfrac{p_{need}}{2}\ \text{(round down)} \\ p_{down} = \dfrac{p_{need}}{2}\ \text{(round up)} \end{cases} \qquad (7)$$

, respectively. Therefore, Theorem 1 is established.

Through the gating of attention weight matrix $A$, we get the output feature maps

$$m_l = C \otimes f\left(A^l + b\right) \quad (l = 1, 2, \cdots, t) \tag{8}$$

where $m_l$, $b \in \mathbb{R}^{(n-h+1) \times 1}$, term $b$ is a bias matrix, $t$ is the number of convolution kernels we use in the attention-gated layer.

We use kernels with different window size $k$ to extract different grained attention weight matrix $A^l$ $(l = 1, 2, \cdots, t)$.

Finally, we get the output feature map

$$M = [m_1, m_2, \cdots, m_t]^T \tag{9}$$

where $M \in \mathbb{R}^{t \times (n-h+1)}$.

We then apply a max-over-time pooling operation [6] over each feature map $M$ to obtain each output $O \in \mathbb{R}^t$ to capture the most important abstract features corresponding to multi-grained attentions, and concatenate all the outputs. These features form the penultimate layer and are passed to the dropout layer [43], then to the fully connected softmax layer.

The detailed process of our model is illustrated in Algorithm A1 (see Appendix A). In some of the model variants, we experiment with "two" and "three" channels of word vectors (see section 4.2), i.e., one channel is fine-tuned via backpropagation while the other channels kept static. We refer to the multi-channel feature of image color in computer vision, and let the convolution kernel $W \in \mathbb{R}^{h \times d \times 2}$ or $W \in \mathbb{R}^{h \times d \times 3}$ if the input has multiple channels. The model is otherwise the same as the single channel model.

*3.1. NLReLU activation function*

Activation functions play a crucial role in achieving remarkable performance in deep neural networks, and ReLU [38] is the most well-known one. The advantage of ReLU is that the gradient is not saturated, thereby avoiding the gradient vanishing problem. ReLU's sparse activation property reduces the interdependence of parameters and also leads to some mathematical advantages [12].

However, the process of training for deep neural networks is complicated by the fact that the distribution of each layer's inputs changes during training, as the parameters of the previous layers change [20]. ReLU does not compress the magnitude of the output data of each neuron, so if the magnitude of the data continues to expand, the deeper the layer of the model, the higher the expansion on the magnitude. Therefore, it is necessary to normalize the data of each layer's inputs to control the distribution of each layer of neuron activations within a reasonable range. To address these problems, Klambauer et al. [26] proposes the SELU activation function, which makes the activation function carry the property of self-normalization on neuron activations.

In this paper, we propose an activation function named Natural Logarithm rescaled ReLU (NLReLU), which is defined as

$$f(x) = \begin{cases} \ln(x + 1.0) & , x > 0 \\ 0 & , \text{otherwise} \end{cases} \tag{10}$$

We use the natural logarithmic transformation to rescale the magnitude of ReLU's positive-axis partial function. The advantages of introducing log transformation into ReLU are as follows:

- **Reduce the heteroscedasticity of the neuron activations' data distribution between the layers of the network.** Log transformation is often used to transform the skewed data before further analyzing [9]. With remedial measures like log transformation reducing heteroscedasticity, raw data from the real world could be analyzed with conventional parametric analysis. We have found that the distribution of neuron activations between each layer of the network is prone to be heteroscedastic. As shown in Fig. 2, we simulate the case where the network has large heteroscedasticity between the layers, and it can be seen that NLReLU transforms each layer's neuron activations to approximately "normal" and effectively reduces heteroscedasticity.

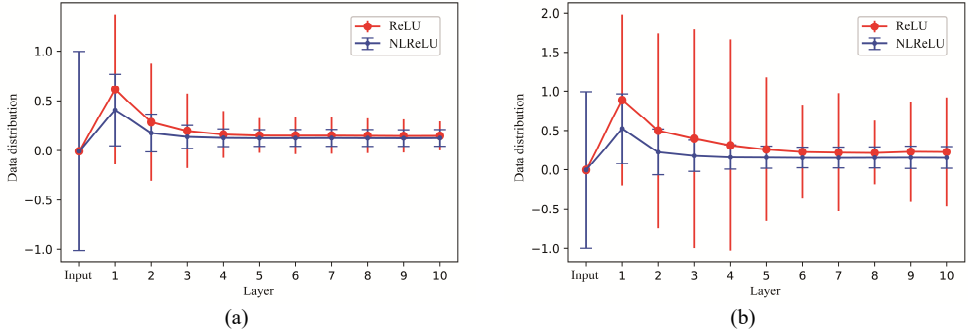

(a)  (b)

**Fig. 2.** Simulation on a fully connected neural network with 10 hidden layers and 100 nodes per hidden layer. The batch size of input data is 100. The inputs obey the standard normal distribution $N(0,1)$, the bias term of the activation functions is initialized as 0.1, and the weights are initialized as: (a) Standard normal distribution $N(0,1)$; (b) Normal distribution $N(0,1.5)$, we simulate the case where the network has large heteroscedasticity between the layers. We recorded the mean and variance of the neuron activations of each layer when using different activation functions. It can be seen that NLReLU could effectively control the distribution of neuron activations and reduce heteroscedasticity between the layers.

- **Control the distribution of neuron activations within a reasonable range.** The natural logarithm function is monotonically increasing in its domain. After taking the log transformation, the relative relationships between the neuron activations are unchanged. But meanwhile, NLReLU compresses the scale in which the data are measured to ensure the output neuron activations of each layer is more stable (see Fig. 2).
- **Make the model more sensitive to differences in majority small value neuron activations.** The property of the natural logarithm function is that the smaller the value of the independent variable, the faster the change in the value decrease of the function. For example, $1.5 - 1.2 = 1.8 - 1.5$ while $\ln(1.5) - \ln(1.2) > \ln(1.8) - \ln(1.5)$; that is, the log transformation makes NLReLU more sensitive to differences in majority small value neuron activations than the larger. The smaller the value of neuron activation, its discrimination is more visible, thus the effect of some larger values is weakened (e.g., noise data).

## 4. Experiments and discussions

Experiments are conducted based on six sentence classification tasks. Section 4.1 introduces the datasets and the comparison systems. Section 4.2 describes the parameter settings and the variants of our model. In Section 4.3, we evaluate the performance of our proposed model on the six benchmark datasets by comparing with other strong baseline models, and analysis the effectiveness of our model. We visualize the feature maps in Section 4.4 to see what information are the model's attention on when making predictions and explore the effects of different activation functions on the performance of our model in Section 4.5.

*4.1. Datasets and comparison systems*

The benchmark datasets used in our experiments are as follows:

- **CR**[1]: This dataset contains customer reviews of various products, e.g., MP3s, cameras and DVD players. The task on this dataset is to predict whether a review is positive or negative [16].
- **MR**[2]: Movie reviews dataset with one sentence per review. Classification involves predicting positive/negative reviews [40].
- **Subj**[3]: Subjectivity dataset where gives the snippets of movie reviews and plot summaries for movies from the internet. Its task is to classify a sentence as being subjective or objective [39].
- **SST-1**[4]: Stanford Sentiment Treebank dataset, an extension of MR but with train/dev/test splits provided and fine-grained labels (including very positive, positive, neutral, negative, very negative) [45].
- **SST-2**: This dataset is derived from SST-1 but removes neutral reviews and converts to two labels, positive and negative, respectively.
- **TREC**[5]: Question dataset, its task involves classifying a question into 6 question types (whether the question is about the person, location, numeric information) [29].

The data preprocessing steps we use is consistent with Kim [24][6]. The statistical summary of these datasets is in Table 1.

We evaluate and compare our model with 13 strong baseline models including:

- SVM$_S$ [42], NBSVM [49] and MNB [49]: Traditional methods of using features such as n-gram as input. Word vectors are not used in these methods.
- Paragraph-Vector [27]: It learns embedding representation of sentences, paragraphs and documents through unsupervised learning based on word vectors and Continuous Bag-Of-Word (CBOW)/Skip-grams model.
- LSTM-based models such as Tree structured LSTM (Tree-LSTM, a generalization of LSTMs to tree-structured network topologies) [48],

---
[1] https://www.cs.uic.edu/~liub/FBS/sentiment-analysis.html#datasets
[2] https://www.cs.cornell.edu/people/pabo/movie-review-data/
[3] http://www.cs.cornell.edu/people/pabo/movie-review-data/
[4] https://nlp.stanford.edu/sentiment/
[5] http://cogcomp.org/Data/QA/QC/
[6] https://github.com/yoonkim/CNN_sentence/blob/master/process_data.py

- Linguistically Regularized LSTM (LR-LSTM) [41], multi-task learning and fine-tuning on LSTM [30] and Dynamic Compositional over Tree structured LSTM (DC-TreeLSTM) [31].
- CNN-based models like Dynamic Convolutional Neural Network with k-max pooling (DCNN) [3], standard CNN for sentence classification (variants including CNN-rand, CNN-static, CNN-non-static and CNN-multichannel) [24], Character-Level CNN (CL-CNN, convolution directly at the character level embedding representation) [54] and Very Deep CNN (VD-CNN, also operating at the character level, but the network architecture goes deeper) [4].
- Capsule network [53]: It explores capsule networks for sentence classification, which uses dynamic routing strategy instead of pooling operation. A standard n-gram convolutional layer is used at the input layer.

**Table 1.** The statistical summary of the datasets after tokenization. *c*: Number of target categories. *l*: Average length of sentence. *N*: Size of dataset. $|V|$: Vocabulary size. $|V_{pre}|$: Number of words present both in $|V|$ and in the set of pre-trained word vectors. *T*: Size of test size (CV means there was no standard training/test split and thus 10-fold Cross-Validation method is used).

| **Dataset** | *c* | *l* | *N* | $|V|$ | $|V_{pre}|$ | *T* |
|---|---|---|---|---|---|---|
| CR | 2 | 19 | 3775 | 5552 | 5053 | CV |
| MR | 2 | 20 | 10662 | 18765 | 16488 | CV |
| Subj | 2 | 23 | 10000 | 21323 | 17913 | CV |
| SST-1 | 5 | 18 | 11855 | 17836 | 16262 | 2210 |
| SST-2 | 2 | 19 | 9613 | 16185 | 14825 | 1821 |
| TREC | 6 | 10 | 5952 | 9493 | 9035 | 500 |

*4.2. Parameter settings and model variations*

We use the publicly available *word2vec*[7] vectors [32,36], that were trained on 100 billion words from Google News, as the pre-trained vectors. The dimensionality of all the input word vectors is 300. Words not present in the set of pre-trained words vocabulary are initialized randomly. For the case of random initializing word vectors, we initialize them to the normal distribution with the mean and variance same as each dataset's vocabulary distribution.

The window sizes of the first convolutional layer's kernels ($h$) are 1, 2, 3, 4, 5 with 100 different kernels each window size, and each window size corresponds to a set of 3 kernels (in the attention-gated layer) with window sizes ($k$) of 1, 3, and 5, respectively. For regularization method we only employ dropout [43] on the penultimate layer with dropout rate of 0.5. The mini-batch size is 50. We choose SELU [26] and NLReLU as the activation function. The parameters of all convolutional layers of the network are initialized with He initialization [17], and the fully connected layer is initialized with Xavier initialization [11].

All the hyper-parameters above are chosen via a coarse grid search on the SST-1's validation set and based on a sensitivity analysis on activation function, and are applied to all datasets. We do not otherwise perform any dataset-specific tuning other than early stopping and learning rate decay. For datasets without a standard validation set we randomly held out 10% of the training data as the validation set. Training is done through stochastic gradient descent over shuffled mini-batches with the Adam update rule [25].

---
[7] https://code.google.com/archive/p/word2vec/

Similar to the standard CNN model [24], we also experiment with several model variants (e.g., AGCNN-rand, AGCNN-static, AGCNN-non-static, and AGCNN-multichannel). Model variations are summarized in Table 2. Unlike the standard CNN model, for the model variant with multiple channels of the input word vectors, the convolution kernels of the first convolutional layer also expand into multiple channels.

Table 2. Summary of model variations.

| Model variant | Description |
| --- | --- |
| AGCNN-rand | All the word vectors are initialized randomly and updated during training. |
| AGCNN-static | This model uses pre-trained word vectors and randomly initializes the unknown words. All words are kept static during training. |
| AGCNN-non-static | Same as the AGCNN-static model but all the word vectors are fine-tuned for each task. |
| AGCNN-multichannel | One channel is fine-tuned during training while the other channels are kept static. All channels are initialized with pre-trained word vectors (random initialization for unknown words). |

*4.3. Performance comparison*

Comparison results of our model against other models are in Table 3. The results show that our model could achieve a general improvement on accuracy highest up to 3.1% (compared with standard CNN models), and gain competitive results over the 13 strong baseline models on four out of the six datasets. Separately speaking, AGCNN-rand improves ranging from 1.0% to 2.5% compared with CNN-rand (despite poor performance on SST-1); AGCNN-static improves ranging from 0.4% to 2.8% compared with CNN-static; AGCNN-non-static improves ranging from 0.1% to 1.6% compared with CNN-non-static; AGCNN-multichannel improves ranging from 0.4% to 3.1% compared with CNN-multichannel (despite poor performance on SST-2). Besides, the performance of our model using NLReLU and using SELU is comparable.

For the binary classification tasks (CR, MR, Subj, and SST-2), a single channel model tends to yield better results when using SELU, while NLReLU is multiple channels. Conversely, for multiple classification tasks (SST-1 and TREC), a model with multiple channels tends to achieve better results when using SELU, while NLReLU is single channel.

On the TREC dataset, the traditional method $SVM_S$ can achieve very high accuracy, and other strong baseline models (e.g., CNN-based and LSTM-based methods) do not exceed this performance. All the performances of our model variants on TREC are close to the accuracy of 95.0% (except the AGCNN-rand), and the best result is slightly higher than $SVM_S$ by 0.3%. Through the comparison of several LSTM-based models, we can see that multi-task learning brings much improvement to LSTM.

Table 4 summaries the comparison results of CNN-static and AGCNN-ReLU-static. It can be seen that in the case of using the same activation function as CNN-static, our model could achieve accuracy improvements ranging from 0.4% to 2.2%.

We can conclude from the Table 3 and Table 4 that our proposal could achieve an effective general improvement over the standard CNN model and is comparable to other strong baseline models. We believe that it is the attention-gated layer helps CNN learn more meaningful and comprehensive text representation and extract the most significant information contained in the sentence. For example, a combination of N-gram convolutional layer with kernel windows of {3,4,5} for a standard CNN model can capture the 3/4/5-gram features of the text. However, for our model, not only the first convolutional layer can capture the 1/2/3/4/5-gram features of the text, but also each feature map is fed into an attention-gated layer with kernel windows of {1,3,5} (from multiple granularities) to enhance the impact of crucial information and suppress the impact of unimportant information.

Table 3. Classification accuracy results of our model against other strong baseline models.

| Model | CR | MR | Subj | SST-1 | SST-2 | TREC |
|---|---|---|---|---|---|---|
| CNN-rand | 79.8 | 76.1 | 89.6 | 45.0 | 82.7 | 91.2 |
| CNN-static | 84.7 | 81.0 | 93.0 | 45.5 | 86.8 | 92.8 |
| CNN-non-static | 84.3 | **81.5** | 93.4 | **48.0** | 87.2 | **93.6** |
| CNN-multichannel | **85.0** | 81.1 | 93.2 | 47.4 | **88.1** | 92.2 |
| SVM$_S$ | — | — | — | — | — | 95.0 |
| NBSVM | 81.8 | 79.4 | 93.2 | — | — | — |
| MNB | 80.0 | 79.0 | 93.6 | — | — | — |
| Paragraph-Vector | 78.1 | 74.8 | 90.5 | 48.7 | 87.8 | 91.8 |
| DCNN | — | — | — | 48.5 | 86.8 | 93.0 |
| CL-CNN | — | — | 88.4 | — | — | 85.7 |
| Tree-LSTM | 83.2 | 80.7 | 91.3 | 48.4 | 85.7 | 91.8 |
| LR-LSTM | 82.5 | 81.5 | 89.9 | 48.2 | 87.5 | — |
| LSTM-Multi-Task (Fine Tuning) | — | — | 94.1 | 49.6 | 87.9 | — |
| VD-CNN | — | — | 88.2 | — | — | 85.4 |
| DC-TreeLSTM | — | 81.7 | 93.7 | — | 87.8 | 93.8 |
| Capsule-Network | 85.1 | 82.3 | 93.8 | — | 86.8 | 92.8 |
| AGCNN-SELU-rand | 82.3 | 78.3 | 91.8 | 44.7 | 83.7 | 92.9 |
| AGCNN-SELU-static | **86.4** | **81.7** | 93.9 | 48.3 | 87.1 | 94.3 |
| AGCNN-SELU-non-static | 85.8 | 81.6 | **94.1** | 49.2 | 87.2 | 94.5 |
| AGCNN-SELU-2-channal | 86.2 | 81.6 | 94.0 | 49.0 | 87.4 | 94.7 |
| AGCNN-SELU-3-channal | 86.1 | 81.3 | 93.5 | 49.4 | 87.6 | 95.3 |
| AGCNN-NLReLU-rand | 82.1 | 78.3 | 91.6 | 44.4 | 83.5 | 92.5 |
| AGCNN-NLReLU-static | **86.2** | 81.6 | 93.8 | 48.0 | 87.2 | 94.3 |
| AGCNN-NLReLU-non-static | 85.0 | 81.3 | 93.9 | **49.6** | 87.3 | **94.9** |
| AGCNN-NLReLU-2-channal | 85.3 | 81.7 | **94.0** | 49.4 | 87.0 | 94.3 |
| AGCNN-NLReLU-3-channal | 85.4 | **81.9** | 93.9 | 49.4 | **87.4** | 94.2 |

Table 4. Performance comparison results (taking the CNN-static and AGCNN-ReLU-static model as an example).

| Model | CR | MR | Subj | SST-1 | SST-2 | TREC |
|---|---|---|---|---|---|---|
| CNN-static | 84.7 | 81.0 | 93.0 | 45.5 | **86.8** | 92.8 |
| AGCNN-ReLU-static | **85.8** | **81.4** | **93.7** | **47.7** | 86.5 | **94.5** |

*4.4. Attention visualization*

In this section, we visualize the feature maps of the standard CNN model and our model (with the same input) as heat maps to compare what features are the model's attention on when making predictions.

The parameters of the models are set as follows: we use a CNN-static model has kernel window size of 1 with 100 feature maps, and ReLU activation; the AGCNN-static model we use has kernel window size of 1 (first convolutional layer) with 100 feature maps and the kernel window size of attention-gated layer is {1,3}, and with NLReLU activation. To rule out the interference of word vector initialization, we use exactly the same initialized word vectors for both models. After training on the MR dataset and saving the models, we randomly select one sentence from the MR as input enter into the two models, the predictions for both models (which is "negative") are correct.

The input sentence is "A turgid little history lesson, humorless and dull" and its category is "negative". We randomly select ten feature maps of CNN-static that are visualized before and after the activation of ReLU (see Fig. 3).

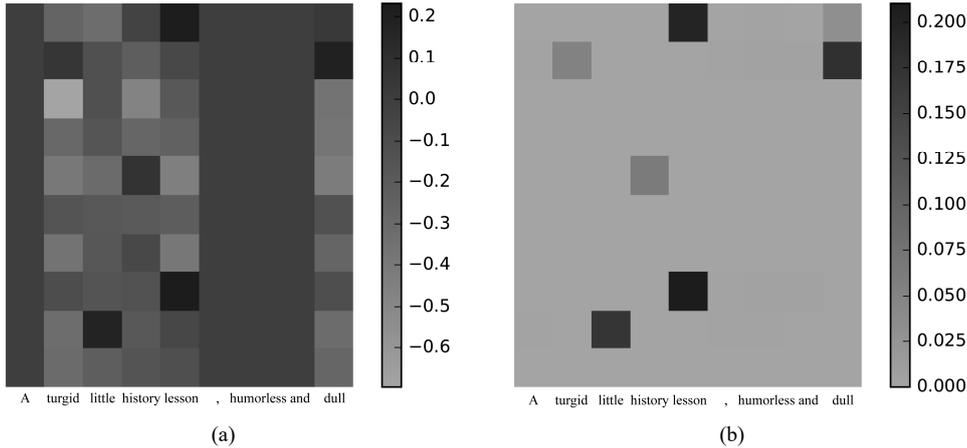

**Fig. 3.** Attention visualization of the CNN-static model: (a) Heat map of the feature maps before activation; (b) Heat map of the feature maps after activation.

As shown in Fig. 3(b), the feature maps are sparse after the activation of ReLU. According to the original intention of the standard CNN model, after the pooling scheme, critical features will be extracted to influence the prediction of the classification. However, it can be seen that the standard CNN model is not ideal for the acquisition of critical features, because words like "turgid", "little", "humorless" and "dull", which all express negative sentiment, are not adequately recognized.

We randomly selected 10 feature maps of the first convolutional layer of AGCNN-static, and their corresponding multi-grained attention feature maps before and after the gating of attention weight matrix are visualized as heat maps, as shown in Fig. 4.

Fig. 4(a) and Fig. 4(b) show that the first convolutional layer of AGCNN-static could extract more distinct and distinguishing features. As depicted in Fig. 4(b) and Fig. 4(d), the convolution in the attention-gated layer of a kernel with window size 3 suppresses the attention to unimportant words' features (e.g., "a", "history", "lesson", and "and"). Besides, the attention of the critical features in Fig. 4(b) are enhanced when compared with other unimportant features (see in Fig. 4(e) and Fig. 4(f)).

The above is a case study of the variation process of features in neural networks. Through intuitive visualization and comparative analysis, we believe that the attention-gated layer could help CNN learn the ability to obtain more comprehensive and meaningful abstract features of the input text, and extract the most significant information contained in the sentence.

*4.5. Sensitivity analysis of activation function*

In this section, we analyze the effects of activation function choices on the model performance. We consider eight different activation functions, including: Sigmoid [10], ReLU [38], Softplus [38], LReLU [34], PReLU [17], ELU [5], SELU [26], and NLReLU. We summary the classification results achieved by AGCNN-static using different activation functions and report them in Table 5.

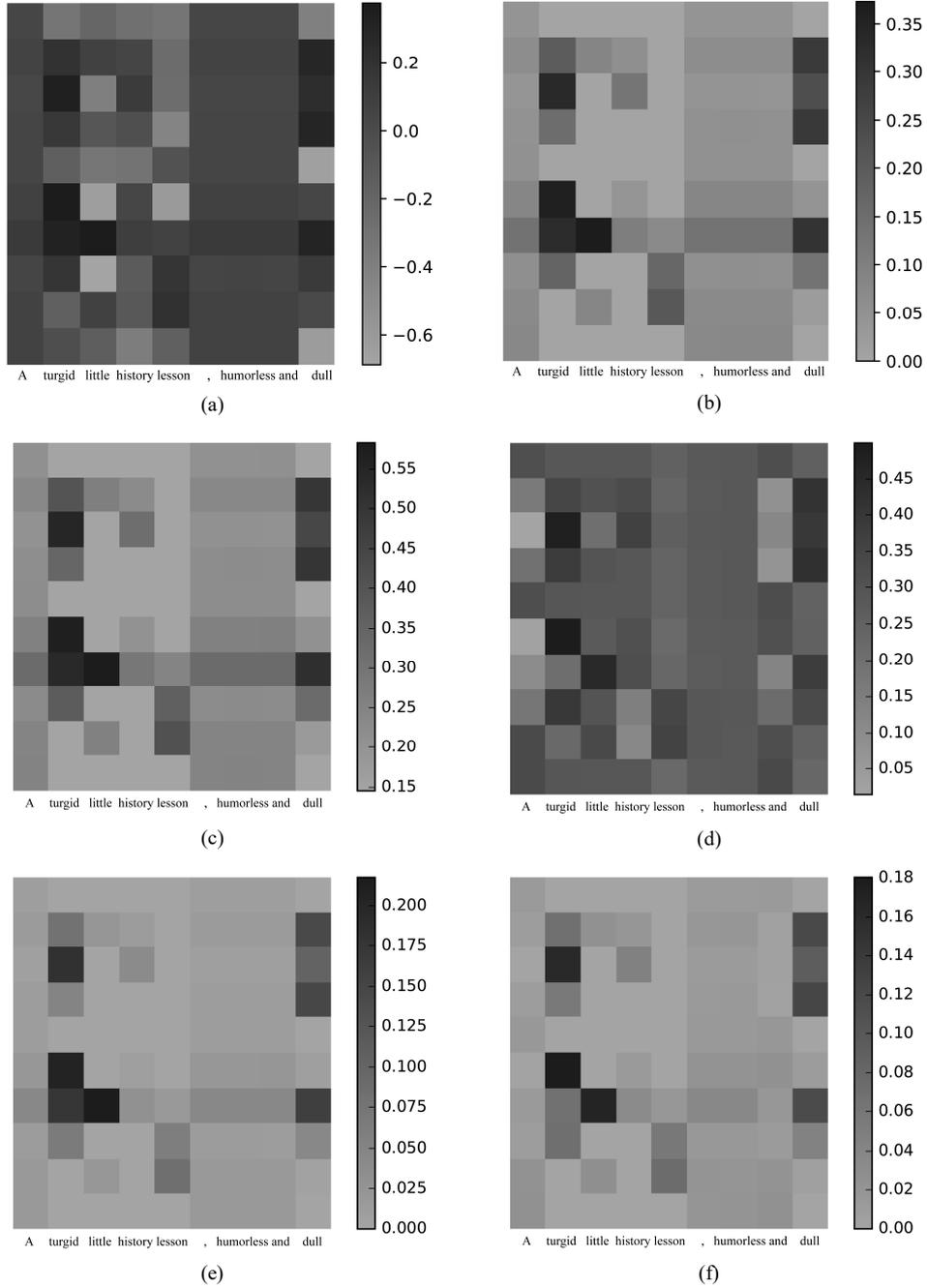

**Fig. 4.** Attention visualization of the AGCNN-static model: (a) Feature maps of the first convolutional layer before the activation of NLReLU; (b) Feature maps of the first convolutional layer after the activation of NLReLU. (c) Feature maps of the attention-gated layer's kernel of window size 1 (after the activation of NLReLU); (d) Feature maps of the attention-gated layer's kernel of window size 3 (after the activation of NLReLU); (e) The element-wise product between the attention weight matrix (c) and the feature map in (b); (f) The element-wise product between the attention weight matrix (d) and the feature map in (b).

On the whole, the best performing activation function is SELU, followed by NLReLU and ELU. One interesting thing is that SELU is obtained by adding self-normalization property to the ELU, and both ELU and SELU have performed well. ReLU could surpass other alternative activation functions on TREC only. Softplus is a smooth approximation to the ReLU, but the log transformation of the exponential function does not bring much performance improvement compared to ReLU. Sigmoid is the worst performer. A very undesirable property of the sigmoid neuron, which is also responsible for poor performance, is that when the neuron's activation saturates at either tail of 0 or 1, the gradient at these regions is almost zero. It is notable that the better performing activation functions tend to transform each layer's neuron activations to approximately "normal" or adequately control the distribution of each layer's distribution of neuron activations. Overall, Table 5 depicts that the performance of NLReLU could outperform ReLU and is comparable to other activation functions.

**Table 5.** Performance comparison experiment results of different activation functions (based on the AGCNN-static model).

| Dataset | ReLU | Softplus | Sigmoid | ELU | PReLU | LReLU | NLReLU | SELU |
|---|---|---|---|---|---|---|---|---|
| CR | 85.8 | 85.7 | 85.5 | 86.0 | 85.7 | 85.9 | 85.8 | **86.4** |
| MR | 81.4 | 81.2 | 81.2 | **81.7** | 81.5 | 81.5 | 81.6 | **81.7** |
| Subj | 93.7 | 93.5 | 93.1 | 93.7 | 93.4 | 93.5 | 93.8 | **93.9** |
| SST-1 | 47.7 | 47.4 | 47.4 | 47.9 | 47.5 | 47.4 | 48.0 | **48.3** |
| SST-2 | 86.5 | 86.2 | 85.2 | 86.6 | 86.6 | 86.5 | **87.2** | 87.1 |
| TREC | **94.5** | 94.3 | 93.8 | 94.2 | 94.1 | 94.1 | 94.3 | 94.3 |

## 5. Conclusions

In this paper, we introduced an attention-gated layer that could help CNN learn more meaningful and comprehensive features, and make full use of limited contextual information in sentences to extract and focus more attention on critical abstract features. We successfully developed the AGCNN model for sentence classification tasks. Experiments on performance comparison and attention visualization of feature maps demonstrated the effectiveness of our proposals and showed that our model can achieve up to 3.1% higher accuracy than the standard CNN models, and gain competitive results over the baselines on four out of the six sentence classification tasks.

We also proposed the NLReLU activation function. We elaborately illustrated and analyzed the advantages of using NLReLU in neural networks. Experiments showed that NLReLU could outperform ReLU and is comparable to other widely-used activation functions on the AGCNN model. For future work, we will explore the application of tree structure and multi-task learning in AGCNN, as well as a more in-depth discussion of the attention gating mechanism and NLReLU.

**Acknowledgments**

We would like to thank the anonymous reviewers for their valuable comments. This work was partially supported by the Foundation for Innovative Research Groups

## Appendix A

**Algorithm A1: AGCNN for sentence classification**

**Input:** The embedding matrix $W_{pre}$ of all pre-trained word vectors for each word in the vocabulary $|V_{pre}|$; the word sequence of sentence $S$, e.g., $S = \{w_1, \cdots, w_n\}$; window size list $l_1$ and $l_2$ of convolution kernels, e.g., $l_1 = \{h_1, h_2, \cdots\}$ and $l_2 = \{k_1, k_2, \cdots\}$; the number of convolution kernels corresponding to each window size in $l_1$ of the first convolutional layer, e.g., $t_1$.

**Output:** The class label of sentence $S$.

1: Initialize the embedding matrix $W_{unknown}$ of all the unknown words, which present in $|V|$ but not in $|V_{pre}|$. Initialize all the weight matrix $W$ and $V$, and bias term $b$ and $D$ in the network.
2: $i = 0$
3: **while** $i < n$ **do**
4:     Look up for $w_i$'s embedding representation $e_i$ in matrix $W_{pre}$ and $W_{unknown}$.
5:     $i = i + 1$
6: **end while**
7: **for** $h$ in $l_1$ **do**
8:     Get the activated feature map $C^h$ after the convolution on $E_{1:n} = [e_1, e_2, \cdots, e_n]^T$ using $t_1$ kernels $W$ with the window size of $h$. (No padding on the input, the stride is 1, and $C^h = [C_1, C_2, \cdots, C_{t_1}] \in \mathbb{R}^{(n-h+1) \times t_1}$)
9:     **for** $k$ in $l_2$ **do**
10:         Get the activated feature map $D^k$ after the convolution on $C^h$ using kernel $V$ with the window size of $k$. (Padding on the input to make the output $D^k$ consistent with the size of the input $C^h$, $D^k \in \mathbb{R}^{(n-h+1) \times t_1}$)
11:         Calculate the element-wise product $m_{h,k} = C^h \otimes D^k$.
12:         Max-over-time pooling on $m_{h,k}$ using a window with a height of $(n-h+1)$ and width of 1 to get the output $m'_{h,k} \in \mathbb{R}^{1 \times t_1}$.
13:     **end for**
14: **end for**
15: Feed the concatenated output $[\cdots, m'_{h_x,k_y}, \cdots]$ ($x, y = 1, 2, \cdots$) to the dropout layer, then to the fully connected softmax layer to produce the prediction $p$.
16: **return** The position of the max value in $p$.